# AMHARIC LLAMA AND LLAVA: MULTIMODAL LLMS FOR LOW RESOURCE LANGUAGES

**Michael Andersland**[*]
mike.andersland@gmail.com

## ABSTRACT

Large Language Models (LLMs) like GPT-4 and LLaMA have shown incredible proficiency at natural language processing tasks and have even begun to excel at tasks across other modalities such as vision and audio. Despite their success, LLMs often struggle to perform well on low-resource languages because there is so little training data available. This shortcoming is especially prevalent with open source models. In this work, we explore training LLaMA-2 to speak Amharic, a language which is spoken by over 50 million people world wide, but has orders of magnitude less data available than languages like English. We employ methods previously used for training LLMs on other languages with data scarcity, and use open source translation models to perform data augmentation and grow our dataset from millions of tokens to billions. We further enhance the capabilities of our model by connecting an image encoder and training on a translated visual instruction tuning dataset in the same manner as LLaVA, resulting in a multimodal Amharic LLM that can understand images along with text. We introduce an Amharic version of a popular benchmarking dataset to evaluate our work. Our models and datasets are open source and available on GitHub.[1].

## 1 INTRODUCTION

The field of natural language processing (NLP) has seen a massive transformation in recent years, spurred primarily by the development of the transformer architecture (Vaswani et al., 2017) and its subsequent application to language modeling via large self supervised neural networks, often with tens or hundreds of billions of parameters trained on trillions of tokens. Recent prominent models include the GPT series (Brown et al., 2020), PaLM (Chowdhery et al., 2022), and LLaMA (Touvron et al., 2023), among others. While state of the art performance is still generally achieved only by proprietary models as of late 2023, the release of LLaMA has given way to a vibrant open source community that has produced near-SOTA open models like Mistral (Jiang et al., 2023) and Mixtral (Jiang et al., 2024). Though LLaMA can only process text inputs, recent projects have added multimodal capabilities by aligning it with pretrained vision or audio encoders (Zhang et al., 2023).

One of the most valuable aspects of these models is their ability to perform few-shot or zero-shot adaptation to novel tasks and instructions without the need for additional training. This is enabled by the massive scale of model parameters and training data. For languages like English, there is an abundance of public data on the internet, enough to provide trillions of tokens for pretraining. For low resource languages that lack high quantities of data, most LLMs either fail to perform tasks at a high level, or do not understand the language at all.

In this work, we focus on developing an open source multimodal language model that can perform NLP tasks in Amharic and understand images. Amharic, the official language of Ethiopia, is a semitic language spoken by over 50 million people worldwide. However, it is vanishingly rare on the public internet. Several Amharic NLP datasets exist, but do not typically exceed 1 million tokens, and are often focused on specific tasks (Tonja et al., 2023). Less than 0.1% of CommonCrawl[2] is Amharic, and even when combining open source datasets without deduplication, we find that less than 500 million tokens of Amharic are available. In addition, the content of this data tends to be biased towards a limited set of topics like news and politics.

---

[*]Work performed while at Garri Logistcs.
[1]https://github.com/iocuydi/amharic-llama-llava

Previous work on language modelling for less represented languages has shown promising results on Chinese (Cui et al., 2023) and Finnish (Luukkonen et al., 2023), with varied approaches including extended pretraining of open source LLMs, development of new pretrained models from scratch, and tokenizer extension. The datasets used to train the models for Chinese and Finnish had tens and hundreds of billions of tokens, respectively. We base our work on the approach used for Chinese Llama and Alpaca, and continue the pretraining of LLaMA-2 with an extended tokenizer.

Even compared to other lower resource languages, the amount of Amharic data available is especially small, with orders of magnitude less data than the prior work. To increase the size and diversity of our data, we apply machine translation to create billions of diverse synthetic Amharic tokens from English text in the RedPajama dataset (Computer, 2023). Until recently, open source translation models were not accurate for Amharic, but work such as Seamless M4T (Barrault et al., 2023) has yielded results approaching and in some cases exceeding the accuracy of proprietary translation models.

Data generation via translation has been effective for translation tasks (Sawai et al., 2021), and we apply this technique with the expectation that the quality of Amharic spoken by the model will not exceed the translation quality, but can still offer an improvement over the small dataset for general NLP and image understanding tasks.

After pretraining on this augmented dataset, we train a small MLP projection (Liu et al., 2023a) to connect a CLIP encoder (Radford et al., 2021) to our pretrained network, and then apply supervised fine tuning on Amharic instruction tuning data (Ouyang et al., 2022) obtained by translating English instruction tuning datasets in a similar manner. Instruction tuning includes pure text conversation pairs and visual instruction data (Liu et al., 2023b) containing image features from CLIP. We evaluate our models with Amharic-MMLU, our Amharic version of the popular LLM benchmark dataset MMLU (Hendrycks et al., 2021). We apply translation again to create Amharic-MMLU from the standard English MMLU.

## 2 Models

LLaMA-2 is an open source foundational language model that has rivaled the performance of similar proprietary models. LLaVA is an open source multimodal model that adds a CLIP vision encoder to LLaMA and trains end to end in order to align image encodings with LLaMA and enable visual understanding and reasoning. LLaMA-2 has variants with 7B, 13B, and 70B parameters. Chat variants tuned for multi turn dialogue settings are available for each model size.

Due to limited resources we use the 7B standard (not chat-tuned) variant for all experiments. As in the previous work, we extend LLaMA-2 pretraining for one epoch, during which the model is trained via next token prediction on unstructured Amharic text. We align a CLIP encoder with the pretrained model by training a small MLP mapping between the two with a translated image captioning dataset. This is followed by one epoch of fine tuning on our multimodal Amharic instruction dataset.

The LLaMA tokenizer is poorly suited to Amharic data. Because Amharic text is so rare relative to other languages in public text, Amharic characters do not have dedicated tokens as more common languages might. For example in English, a single word might map to a single token. For the Ge'ez characters that make up Amharic (and other rare characters) the LLaMA tokenizer and others solve this issue by representing them with a combination of multiple generic byte tokens. In some cases this means that a single Amharic word may be encoded to 10+ tokens, while an equivalent English word might only require a single token.

---

[2]https://commoncrawl.org/

Table 1: Tokenization Comparison (excluding start and stop tokens)

| Tokenizer | Text | Tokenized Text |
|---|---|---|
| LLaMA | Hi, how are you? | 6324, 29892, 920, 526, 366 |
| Amharic LLaMA | Hi, how are you? | 6324, 29892, 920, 526, 366 |
| LLaMA | ሰላም፥ እንዴት ነህ? | 29871, 228, 139, 179, 228, 139, 142, 228, 139, 160, 228, 144, 166, 29871, 228, 141, 168, 228, 141, 152, 228, 142, 183, 228, 140, 184, 29871, 228, 141, 147, 228, 139, 136 |
| Amharic LLaMA | ሰላም፥ እንዴት ነህ? | 46702, 32547, 35199 |

In addition to impeding learning, this tokenization scheme increases the sequence length for any text being processed, which is problematic for the scaling of the transformer with respect to sequence length, and greatly reduces the effective context window and processing speed (this effect can even be observed on proprietary models, which often stream responses more slowly when queried with rare characters). Following the approach of Chinese LLaMA, we use SentencePiece (Kudo & Richardson, 2018) to learn a tokenization scheme for Amharic from our public (untranslated) Amharic data, resulting in an Amharic token vocabulary of 19008 tokens. We combine this with the LLaMA tokenizer of 32000 tokens for a new vocabulary size of 51008. We leave the original embeddings unchanged, though they can update during training when English tokens occasionally appear in the Amharic data. Table 1 illustrates the improved tokenization with the new vocabulary.

## 3 Data

For our pretraining task, we use a combined dataset consisting of 436 million tokens from public sources including Common Crawl, Azime & Mohammed (2021) and various web scrapes, along with an additional 3.348 billion Amharic tokens translated from the Red-Pajama dataset, specifically the sections containing text from Wikipedia and from various books. Table 2 contains the exact proportions.

Table 2: Composition of our Amharic pretraining dataset

| Data Source | Percentage | Tokens |
|---|---|---|
| Translated RedPajama Wikipedia | 48% | 1826m |
| Translated RedPajama Books | 40% | 1522m |
| Real Amharic Text | 12% | 436m |

To translate the English data, we apply the Seamless M4T large model with text to text translation, specifying English to Amharic. For the image encoder alignment and visual instruction tuning, we use Seamless M4T to translate the BLIP caption dataset (Li et al., 2022) and the LLaVA visual instruction tuning dataset.

The pure text instruction pairs in our fine tuning task include Amharic translated versions of the Alpaca (Taori et al., 2023), Dolly (Conover et al., 2023), and OpenAssistant (Köpf et al., 2023) datasets. We use the Google Translate API to translate Alpaca and Dolly, and Seamless M4T to translate OpenAssistant. We prune the OpenAssistant conversation trees to ensure that only highly rated responses are used in our dataset. In addition to these Amharic datasets, we create mixed English and Amharic datasets to help the model leverage existing knowledge of English tokens to learn more about the new Amharic tokens and their relation to concepts that the model may already understand well in English. We replace either the human or AI role in an instruction pair with the original untranslated data, and add a specification to the prompt indicating the language in which the AI role is expected to answer. We further augment this data with a translation task in which either the human or AI role of a translated instruction pair has its English and Amharic versions inserted into a new synthetic instruction pair specifying a translation.

Our Seamless M4T translations are performed on an A100 GPU over several weeks. Because Seamless M4T performance can suffer with long sequences, the text is translated in chunks of a few sentences at a time, not exceeding a fixed token limit, with very long sentences excluded entirely. To speed up translation we use batch inference and map sentences into different sized batches depending on length, reordering thema after translation.

## 4 Experiments

We follow the experimental setup and hyperparamter configurations used in the 7b parameter model subset of the Chinese LLaMA experiments. For visual instruction tuning, we follow the experimental setup used in LLaVA-1.5. When combining visual and text-only instruction tuning, we use the LLaVA-1.5 setup.

We use LoRA (Hu et al., 2022) for the attention layers but do not train with quantization. We train each model for one epoch on a single A100 GPU, which takes 1-4 weeks for pretraining and 2-7 days for finetuning. Pretraining is more computationally expensive, and with our limited resources we choose to perform most of our experiments at the finetuning stage. We perform one pretraining run with 436m tokens gathered from public sources, and another with our augmented 3784m token dataset including mostly synthetic translated data. We use the 3784m model as a base for most of our finetuning experiments.

We finetune with different versions of our dataset to explore the effects of including English data and the translation task versus pure Amharic data. We compare tuning on just Amharic data, English data followed by Amharic data, and all the data together.

We compare the effect of visual instruction tuning on pure text tasks, and explore the order in which the model is exposed to visual data relative to other training phases. We finetune with visual and text-only instruction data and compare this to omitting text-only data.

## 5 Results and Evaluation

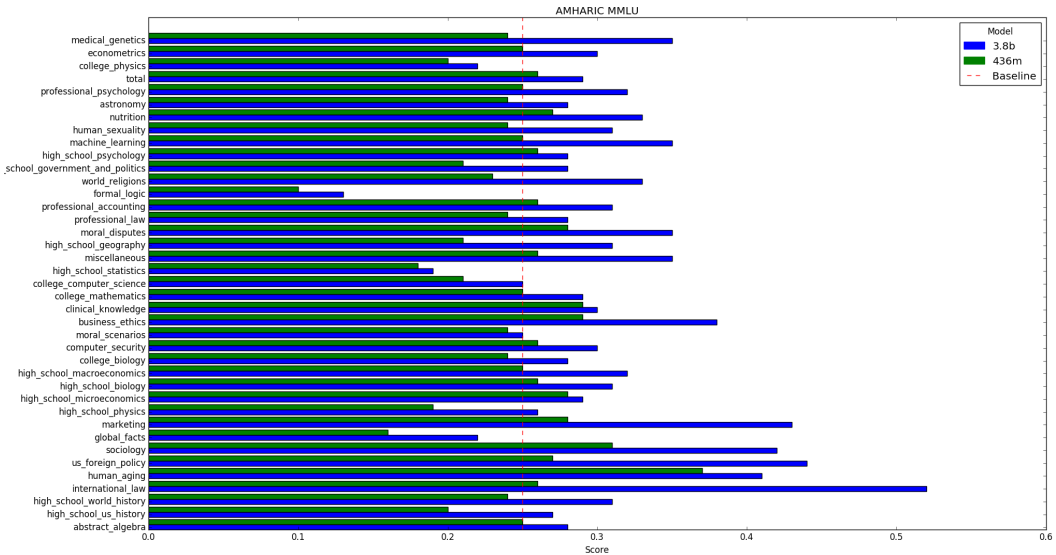

Figure 1: Amharic MMLU Subject Performance, 436m vs 3784m token dataset

Quantitative evaluation is difficult for low resource languages. As with training data, well measured benchmarks and even baselines against which to benchmark are scarce. We use SeamlessM4T again to create a rudimentary Amharic version of the widely used English MMLU dataset for language task evaluation. We test the text understanding and world knowledge of our models with MMLU by asking multiple choice questions. Across most topics, the variants pretrained with the augmented dataset outperform those using the smaller datasets, with some notable exceptions. Both models fail to even exceed a baseline of a random guess on several STEM topics like math, logic, and physics, as shown in Figure 1. We suspect that this may be due to the nature of the questions, where the mistranslation of a single character could completely alter the meaning of the question and the answer. In contrast, the models tend to significantly outperform the baseline on topics like law and ethics. Table 3 compares models across all Amharic MMLU topics, and again with STEM topics excluded. Qualitatively, our models perform well at a variety of tasks including text generation, summarization and named entity recognition, as well as visual question answering and analysis of multiple pictures. Figures 2 and 3 contain examples. Though both visual models were able to provide coherent answers about images, they excel in different areas. The model that saw only visual instruction data and no pure text data tends to give better responses to multimodal queries, but falls short on pure text queries. The model that saw the full extent of visual and text data outperforms all pure text variants and has the best overall performance on Amharic MMLU, but offers less detailed responses to multimodal queries and is more likely to make mistakes about details in images. Both visual models can respond to queries containing multiple images as shown in Figure 4, but performance deteriorates rapidly as the number of images increases, making the models unable to understand video. Further qualitative examples, comparisons to GPT-4 (OpenAI et al., 2023) and failure cases are included in the appendix.

Table 3: Amharic MMLU Scores with and without STEM topics. Sequential refers to splitting the full text dataset and training on the mixed English-Amharic data before any pure Amharic data. Full text refers to the entire finetuning dataset except the multimodal visual examples.

| Pretrain | Finetune | Amharic MMLU Score | Amharic MMLU Non-STEM Score |
|---|---|---|---|
| 436m | Full Text | 0.26 | 0.30 |
| 3784m | Amharic Only | 0.27 | 0.32 |
| 3784m | Sequential | 0.28 | 0.35 |
| 3784m | Full Text | 0.29 | 0.37 |
| 3784m | Full Text + Visual | **0.30** | **0.38** |
| 3784m | Visual Only | 0.26 | 0.27 |

| Prompt | Response |
|---|---|
| ከኢ.ትዮጵያ ወደ አሜሪካ ለሚሄድ ሰው የቱሪስት ቪዛ ማመልከቻ ደብዳቤ ይጻፉ። | ውድ ቪዛ አፊሰር፡ ይህ ደብዳቤ በደንብ እንደሚያገኝዎ ተስፋ አደርጋለሁ። በቅርቡ ወደ አሜሪካ ለመጓዝ እና ለሦስት ወራት ያህል ለመቆየት እንናፍቃለሁ። ለጉዞዬ ቪዛ ለማውጣት ወደ ኤምባሲው ማመልከቻዬን ለማቅረብ እጽፍልሃለሁ። ለጉዞዬ ወጪዎች' ለመሸፈን እና ራሴን ለመደገፍ ወደ አሜሪካ በምሄድበት ወቅት ለመስራት የሚያስችለ' በቂ ገንዘብ እና ንብረት እንዳለኝ ማረጋገጥ እፈልጋለሁ። በተጨማሪም በጉዞዬ ወቅት ምንም አይነት የወንጀል ክስ፣ የጤና ችግር ወይም ተላላፊ በሽታ እንደሌለኝ አረጋግጣለሁ። በጉዞዎቼ ወቅት፣ የአሜሪካን ህግ እና ስርዓት በማክበር እና የአሜሪካን ማህበረሰብ ወክዬ እንድወክል ተመኝሁ። ማመልከቻዬን ስላጤንክኝ አመሰግናለሁ። |
| Write a tourist visa application letter for someone visiting the USA from Ethiopia | Dear Visa Officer, I hope this letter finds you well. I am looking forward to traveling to America soon and staying for three months. I am writing to you to submit my application to the embassy to issue a visa for my trip. I want to make sure I have enough money and assets to work while I'm in America to cover my travel expenses and support myself. I also certify that I have no criminal charges, health problems or communicable diseases during my trip. During my travels, I aspire to uphold American law and order and represent the American community. Thank you for considering my application. |

Figure 2: Example output for a writing task. An approximate translation in English is included for both the prompt and response.

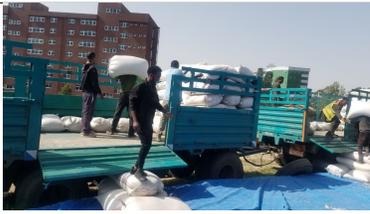

| Prompt | Response |
|--------|----------|
| በዚህ ምስል ውስጥ ምን እየሆነ ነው? | በሥዕሉ ላይ አንድ የሰዎች ቡድን በትልቅ ሰማያዊ የጭነት መኪና ጀርባ ላይ ቆሞ ትልቅ ወይም ከረጢት እየጫነ ነው። የጭነት መኪናው አልጋው ላይ በርካታ ከረጢቶችኖያሉት ሲሆን እንዳንዶቹም በሰዎች እየተጫኑ ነው። የጭነት መኪናውና የሰዎች ቡድን ከቤት ውጭ ያለ ይመስላል ፤ ምናልባትም እንዳ የግንባታ ቦታ ወይም ከቤት ውጭ ክስተት ባለ ቦታዎች ላይ። የጭነት መኪናው እና የሰዎች ቡድን መኖራቸው የጭነት መኪናው ትልቅ ጭነት ለማጓጓዝ ጥቅም ላይ እየዋለ መሆኑን ያመለክታል |
| What is happening in this picture? | In the picture, a group of people is standing on the back of a big blue truck loading a large bag or sack. The truck has several bags on the bed, some of which are being loaded by people. The truck and group of people appear to be outdoors, perhaps at a construction site or outdoor event. The presence of a truck and a group of people indicates that the truck is being used to transport large loads |

Figure 3: Example output for an Amharic multimodal query. An approximate translation to English is included for both the prompt and response.

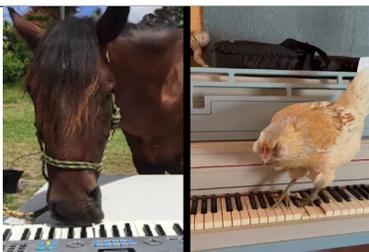

| Prompt | Response |
|--------|----------|
| ሁለቱ ሥዕሎች ምን የሚያመሳስላቸው ነገር አለ? በመካከላቸው ያለው ልዩነት ምንድን ነው? | በሁለቱ ሥዕሎች መካከል ያለው የጋራ ነገር ሁለቱም እንስሳት ፒያኖን እየተጫወቱ መሆናቸው ነው። ይሁን እንጂ በሁለቱ ሥዕሎች መካከል ያለው ልዩነት ፒያኖው ላይ ያለው እንስሳ ነው። አንድ ሥዕል አንድ ፈረስ ፒያኖውን ሲጫወት የሚያሳይ ሲሆን ሌላኛው ደግሞ የዶሮ ፒያኖውን ሲጫወት የሚያሳይ ነው። ሁለቱም እንስሳት ፒያኖውን ለመጫወት ሲሞክሩ የሚያሳዩት ትዕይንት አስደሳችና አዝናኝ ነው። |
| What do the two pictures have in common? What is different between them? | The common denominator between the two pictures is that both animals are playing the piano. However, the difference between the two pictures is the animal on the piano. One picture shows a horse playing the piano and the other shows a chicken playing the piano. The scene of both animals trying to play the piano is hilarious. |

Figure 4: Example output for an Amharic multimodal query. An approximate translation to English is included for both the prompt and response.

## 6  Conclusion

We have explored solutions for improving pretrained language models by adding the ability to process and understand a low resource language. We have presented data augmentation via translation as a way to rapidly grow the available dataset by at least an order of magnitude while significantly increasing the diversity, yielding improved performance on text and visual instruction tasks after additional fine tuning. We applied a similar methodology to create a baseline benchmark and measure the performance of our models.

### 6.1  Limitations

These models have limitations that should be considered when extending or deploying them.

- They may exhibit bias or generate harmful content.

- The models are brittle and can hallucinate or deliver unpredictable outputs, especially with high temperature and sampling.

- Our finetuning data mostly consists of single turn pairs, and performance degrades in conversational chat settings as the dialogue grows.

- Further evaluation and tuning with highly accurate real world Amharic datasets is needed before production deployments.

- Though their understanding of Amharic characters has improved, the models lack cultural knowledge that might be expected to come along with understanding of Amharic, as the majority of the Amharic they saw was synthetic data from translated English.

### 6.2  Future Work

The area of applying large models to low resource languages is ripe for exploration and stands to benefit many populations that cannot make effective use of today's models. Future projects could include:

- Creating high quality human-informed Amharic datasets like AmQA (Abedissa et al., 2023) for fine tuning and evaluation, which can significantly exceed the quality of the translated datasets and may offer outsized gains when applied during the fine tuning step, and more comprehensive and accurate quantitative evaluation.

- Performing additional tuning steps like Reinforcement Learning with Human Feedback (RLHF) to improve performance and alignment .

- Experimenting with LLaMA-2-Chat model variants, other model sizes, quantization, and new architectures like Mixtral.

- Applying these techniques to other poorly represented languages on which today's models struggle.

- Expanding the experiments done in this paper with larger synthetic datasets to learn about how this approach scales.

- Adding Amharic OCR capabilities to the model by procedurally generating images of noisy Amharic text, finetuning CLIP if necessary, and augmenting the visual instruction tuning dataset with LLM-generated instruction sequences that reference the images.

- Improving the efficiency of our translation batching algorithm and the cleaning and management of translated data including removal of poorly translated sentences and sentences translated from unexpected source languages.

- Experimenting with Retrieval Augmented Generation (RAG) (Lewis et al., 2020) and Amharic document embeddings.

- Further enhancing these models with video or audio understanding.

## 6.3 Acknowledgements

Thanks to Daniel Temesgen for insightful discussions on the Amharic language and the challenges of language modelling and data preparation for low resource languages. Thanks to the Google for Startups and Google Cloud teams for assisting with consistent access to A100 GPUs and cloud credits.

## A  APPENDIX

We present additional qualitative examples of model performance with image and text instructions. Figure 5 includes an intentionally confusing scene where the model partially understands but confuses the act of ironing with washing clothes. Both this and Figure 6 include comparisons to GPT-4's Amharic visual reasoning capabilities via the GPT-4-vision-preview API. Amharic LLaVA generally demonstrates improved Amharic visual reasoning over GPT-4, which is impeded by a poor understanding of Amharic. GPT-4's Amharic has improved rapidly since its release, and will likely continue to get better.

Figure 7 shows a longer text generation instruction in which the model provides a mostly valid answer with some erroneous repetition on its suggested third step of the response. Figure 8 shows an example of model shortcomings on Amharic text. The model infers some general meaning of the text from the context, but fails to accurately decipher the text. This is likely due to the rarity of visual Amharic text data in the CLIP training set as well as the complete absence of Amharic text from the images in our visual finetuning data, but may be easily solved with a procedurally generated synthetic Amharic OCR dataset.

Figure 9 shows the model unsuccessfully attempting to interpret a video presented as a series of frames stitched together. The model understands the frames in isolation but is unable to identify the activity performed across them.

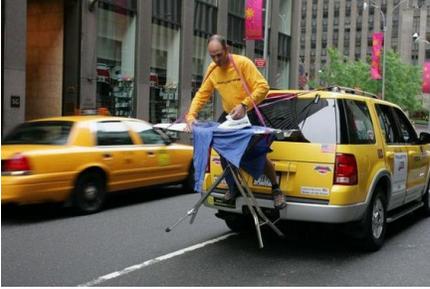

| Model | Prompt | Response |
|-------|--------|----------|
| Amharic-LLaVA | በዚህ ምስል ላይ ያልተለመደው ምንድን ነው? | በዚህ ምስል ውስጥ ያልተለመደ ገጽታ አንድ ሰው ቢጫ ታክሲ ጀርባ ላይ የልብስ ማጠቢያ መስመርን በመጠቀም የልብስ ማጠብያውን ማጠብ ነው። ይህ የተለመደ እይታ አይደለም ፤ ምክንያቱም ሰዎች በተለምዶ የልብስ ማጠቢያ መስመሩን በቤት ውስጥ ወይም በንግድ ተቋማት ውስጥ ይጠቀማሉ። ሰውየው በልብስ ማጠቢያ መስመሩ ላይ ተንጠልጥሎ መገኘቱ ትዕይንቱን አስደሳችና ያልተለመደ ያደርገዋል |
| | What is unusual about this picture? | An unusual aspect in this picture is that a man is washing his clothes using a clothesline on the back of a yellow taxi. This is not a common sight, because people usually use the clothesline at home or in business establishments. The fact that the man is hanging on the clothesline makes the scene interesting and unusual. |
| GPT-4 (2023) | በዚህ ምስል ላይ ያልተለመደው ምንድን ነው? | በምስሉ ላይ የታየውን በመሆኑ ሊመረጥ አይችልም እንጂ እንደ አዲስ ያልታወቀ አስቤ ነገር የሚታየው፣ አንድ ሰው መኪና ተሸክሞው ትንሽ የመጥበሻው ማዕከላዊ ክፈና ላይ ማንጠባጠብ አገናኝ ነው። ይህ የመንገድ ላይ እንደሚከናወን ወይም የሞክር ሥራ ባለመሆኑ እና ሰውች ለመሳቀፍ ይችላል። የተሳሳተ መስሪያ መጠቀም፣ ከመንገድ ላይመኪና አፍነጫ መከታተል፣ ወይም አነስተኛውን ናዳ ለመቁጠር የመጠቀም ስምንት ሊታይ ይችላል። በእውነት ምስሉ አነስተኛውን መንገድ ፉና እስከነው አንድ ብልጊ ሥራ ወይም የአስገራሚ የተንኮል ስጦታ መሆን እና የታዘዘውን ሃላፊነት መቻል ሊኖር ይችላል። |
| | What is unusual about this picture? | It can't be selected because of what's shown in the picture, but what appears to be a new unknown object, a person carrying a car, is a link dripping onto the central frame of the pan. This is done on the street or because it's not a test job and people can get yelled at. Using the wrong device, tracking the car off the road, or using the smallest number to count eight can be seen. Indeed, as long as the picture is taken in the least possible way, it is possible that it is a clever work or a wonderful gift of cunning, and that there can be a responsibility that is commanded. |

Figure 5: Example output for an Amharic multimodal query. An approximate translation to English is included for both the prompt and response.

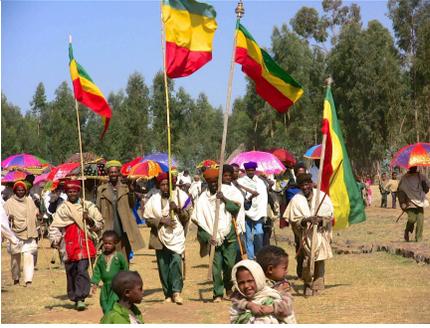

| Model | Prompt | Response |
|---|---|---|
| Amharic-LLaVA | በዚህ ምዕል ላይ ምን እየሆነ ነው? | በም፡ዕሉ ላይ አንድ የሰዎች ቡድን በቀለማት ያሽበረቁ ባንዲራዎችን ይዘው እየተራመዱ ሲሆን አንዳንዶቹም ጃንጥላ ይዘው ነው። በሕዝቡ መካከል አንድ ሕፃን ወይም አንድ ትንሽ ልጅ እየተራመደ ሲሆን ምናልባትም ከቤተሰቡጋር እየተጓዘ ወይም እየተጓዘ ሊሆን ይችላል። ሕዝቡ ባንዲራዎችን እና ጃንጥላዎችን በመጠቀም የበዓል ወይም ሁኔታን ይፈጥራል |
| | What is happening in this picture? | In the picture, a group of people are walking with colorful flags, some of them carrying umbrellas. In the middle of the crowd is walking a baby or a small child, possibly walking or traveling with his family. The people use flags and umbrellas to create a festive atmosphere |
| GPT-4 (2023) | በዚህ ምዕል ላይ ምን እየሆነ ነው? | በተመለከተው ም፡ዕል ላይ እንደምታይበት ሰዎች ከባንዲራቸው ባለቤቶችን ከብርሃን መከላከያ አምሮች ጋር መጓዣጓም ይኖራቸዋል። እርስ በርስ በሚገጣጠሙ ከባንዲራቸው ባለቤቶች እንደ ኢትዮጵያውያን ወንዶች ናቸው። ብዙም ከእነሱ አብራ ከየ ቤልጥና መከላከያ ትግጦኛ መገጊያ አድርገው ተይዘው እንደሚዳሉ። ሰዎች ባለቤቶች የተመነጨ ልብስ የለበሱ የሚታይ ነበር ይበተንም። ምዕሉ በኢትዮጵያ ውስጥ የተወሰደ እንደ መሆኑ አመጣጥ ይችላል፤ የሚታዩት የኢትዮጵያ ባንዲራን ወንበርነት ያለቸውና አንዳንድ የተለያዩ አምሮች በእጁ የያዙ ሰዎችን ማንበብ እንደሚቻል አይታይበታል። ትልቅ ባለቤቶችና ባለውነት የሚገልጹ ከብርሃን መከላከያ አምሮች መጠቀም የሚታይም ነው፤ እስከ ዛሬም ድረስ በማኅበረሰብ ላይ ከወንበሩ ልዩ እንቅስቃሴ እንዳለበት ሊመስል አለበት። |
| | What is happening in this picture? | As you can see in the picture above, people will have to go with their flag holders and light shields. They are Ethiopian men who are the owners of their flags facing each other. That they will be caught in a tight barrier if they are not far away from them. The owners of the people were seen wearing original clothes. It can be said that the picture was taken in Ethiopia. It is not seen that it is possible to read people who are seen with the Ethiopian flag as a chair and some different amours in their hands. The use of light protection amours is seen by large owners and those who express their wealth. To this day, it must seem that he has a special influence on society from his chair. |

Figure 6: Example output for an Amharic multimodal query. An approximate translation to English is included for both the prompt and response.

| Prompt | Response |
|---|---|
| እባክዎን የፒዛ ምግብ ቤት ለመክፈት 7 ምክሮችን ይስጡኝ | 1. ትክክለኛውን ቦታ ይምረጡ፦ የፒዛ ምግብ ቤት በጣም ጥሩ ቦታ ማግኘት አስፈላጊ ነው ምክንያታም ደንበኞችዎ በቀላሉ ማግኘት እና ወደ ሬስቶራንቱ መምጣት አለባቸው። |
| | 2. ጥሩ ድባብ ይፍጠሩ፦ የፒዛ ምግብ ቤት ድባብ ሞቅ ያለ እና እንግዳ ተቀባይ መሆን አለበት ይህም ደንበኞችን እንዲዝናኑ እና እንዲመገቡ ያደርጋል። |
| | 3. ከፍተኛ ጥራት ያላቸውን ንጥረ ነገሮች ተጠቀም፦ የፒዛ ምግብ ቤት ደንበኞችን ለማስደሰት ከፍተኛ ጥራት ያላቸውን ንጥረነገሮች መጠቀም አስፈላጊ ነው። ትኩስ, ትኩስ እና ትኩስ የሆኑ ንጥረ ነገሮችን ይጠቀሙ እና ትኩስ, ትኩስ እና ትኩስ የሆኑ ንጥረ ነገሮችን ይጠቀሙ. |
| | 4. ጥሩ የፒዛ ሼፍ ቀጥል፦ ፒዛ ሼፍ ደንበኞችን ለማስደሰት እና ከፍተኛ ጥራት ያላቸውን ምግቦች ለማቅረብ እውቀት ያለው እና ልምድ ያለው መሆን አለበት። |
| | 5. ጥሩ አገልግሎት ያቅርቡ፦ የፒዛ ምግብ ቤት ሰራተኞች ለደንበኞችዎ ወዳጃዊ እና ቀልጣፋ አገልግሎት መስጠት አለባቸው። |
| | 6. ጥሩ ዋጋዎችን ያቅርቡ፦ የፒዛ ምግብ ቤት ደንበኞችን ለመሳብ እና ለማቆየት ተወዳዳሪ ዋጋዎችን መስጠት አስፈላጊ ነው። |
| | 7. ጥሩ ግብይት እና ማስተዋወቅ፦ የፒዛ ምግብ ቤት ደንበኞችን ለመሳብ እና ለማቆየት ጥሩ የግብይት እና የማስተዋወቂያ ዘመቻዎችን ይጠቀሙ። ይህ ማህበራዊ ሚዲያን፣ የመስመር ላይ ማስታወቂያዎችን እና የሀገር ውስጥ ዝግጅቶችን ሊያካትት ይችላል። |
| Please give me 7 tips to open up a pizza restaurant | 1. Choose the right location: Finding the best location for a pizza restaurant is important because your customers need to find it easily and come to the restaurant. |
| | 2. Create a good atmosphere: The atmosphere of a pizza restaurant should be warm and welcoming, which will make customers feel relaxed and fed. |
| | 3. Use high quality ingredients: In order to please the customers of a pizza restaurant, it is important to use high quality ingredients. Use fresh, fresh and fresh ingredients and use fresh, fresh and fresh ingredients. |
| | 4. Resume of a good pizza chef: A pizza chef must be knowledgeable and experienced to delight customers and serve high-quality dishes. |
| | 5. Provide good service: Pizza restaurant staff should provide friendly and efficient service to your customers. |
| | 6. Offer good prices: To attract and retain pizza restaurant customers, it is important to offer competitive prices. |
| | 7. Good Marketing and Promotion: Use good marketing and promotional campaigns to attract and retain pizza restaurant customers. This may include social media, online advertising and local events. |

Figure 7: Example output for a writing task. An approximate translation in English is included for both the prompt and response.

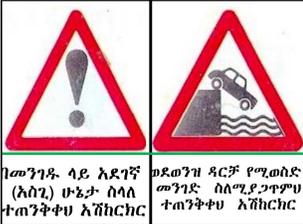

| Prompt | Response |
|---|---|
| ይህ ሥዕል ምንድን ነው? | ምስሉ ሁለት ቀይ እና ነጭ የትራፊክ ምልክቶችን ያሳያል ፤ አንዱ ወደ ግራ እና ሌላኛው ወደ ቀኝ። ሁለቱም ምልክቶች "አደጋ" የሚል ቃል አላቸው ፤ ይህም አሽከርካሪዎች ጥንቃቄ እንዲያደርጉ እና ጥንቃቄ እንዲያደርጉ ማስጠንቀቂያ ነው። " ምልክቶቹ በትራፊክ ምልክቶች ላይ የተለመዱ ቀይ እና ነጭ ቀለሞች አሏቸው ፤ ይህም ለእሽከርካሪዎች እና ለእግረኞች በቀላሉ እንዲታዩ ያደርጋቸዋል |
| What is this picture? | The image shows two red and white traffic signs, one to the left and the other to the right. Both signs have the word "DANGER" on them, which is a warning to drivers to be careful and take care. And it makes them easily visible to pedestrians |

Figure 8: Example output for an Amharic multimodal query. An approximate translation to English is included for both the prompt and response.

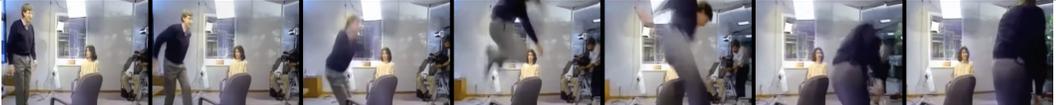

| Prompt | Response |
|---|---|
| በቪዲዮው ውስጥ ምን እየሆነ ነው? | በቪዲዮው ውስጥ አንድ ሰው በአንድ ክፍል ውስጥ በተለያዩ ቦታዎች ላይ ቆሞ በርካታ ፎቶግራፎችን ሲወስድ ይታያል |
| What is happening in the video? | In the video, a person is seen standing in different places in the same room and taking several pictures |

Figure 9: Example output for an Amharic multimodal query. An approximate translation to English is included for both the prompt and response.